\def\year{2022}\relax
\documentclass[letterpaper]{article} 
\usepackage{aaai24}  
\usepackage{times}  
\usepackage{helvet}  
\usepackage{courier}  

\usepackage{times}
\usepackage{latexsym}

\usepackage[hyphens]{url}  
\usepackage{graphicx} 
\usepackage{natbib}  
\usepackage{caption} 
\DeclareCaptionStyle{ruled}{labelfont=normalfont,labelsep=colon,strut=off} 
\usepackage{algorithm}

\usepackage{color}
\usepackage{graphicx}
\usepackage{tabularx}
\usepackage{soul}
\usepackage{booktabs}
\usepackage{algorithm}
\usepackage{algpseudocode}
\usepackage{amsmath}
\usepackage{multirow}
\usepackage{paralist}

\usepackage{xcolor}

\usepackage{newfloat}
\usepackage{listings}
\floatstyle{ruled}
\newfloat{listing}{tb}{lst}{}
\floatname{listing}{Listing}
\title{Detecting harassment and defamation in cyberbullying \\with
emotion-adaptive training}
\author{
    Peiling Yi, Arkaitz Zubiaga, Yunfei Long
     }
\affiliations{
}

\begin{document}
\maketitle
\begin{abstract}
Existing research on detecting cyberbullying incidents on social media has primarily concentrated on harassment and is typically approached as a binary classification task. However, cyberbullying encompasses various forms, such as denigration and harassment, which celebrities frequently face. Furthermore, suitable training data for these diverse forms of cyberbullying remains scarce. In this study, we first develop a celebrity cyberbullying dataset that encompasses two distinct types of incidents: harassment and defamation. We investigate various types of transformer-based models, namely masked (RoBERTa, Bert and DistilBert), replacing (Electra), autoregressive (XLnet), masked\&permuted (Mpnet), text-text (T5) and large language models (Llama2 and Llama3) under low source settings. We find that they perform competitively on explicit harassment binary detection, however, their performance is substantially lower on harassment and denigration multi-classification tasks.  
Therefore, we propose an emotion-adaptive training framework (EAT) that helps transfer knowledge from the domain of emotion detection to the domain of cyberbullying detection to help detect indirect cyberbullying events. EAT consistently improves the average macro F1, precision and recall by 20\% in cyberbullying detection tasks across nine transformer-based models under low-resource settings. Our claims are supported by intuitive theoretical insights and extensive experiments.\footnote{The data and source are publicly available at \url{https://github.com/Misinformation-emotion/Cyberbullying-emotion}}\newline \textcolor{red}{Warning: This paper contains offensive words, which do not reflect the views of the authors}.

 
\end{abstract}

\section{Introduction}
\begin{figure}[tbh]
  \centering
  \includegraphics[width=\linewidth]{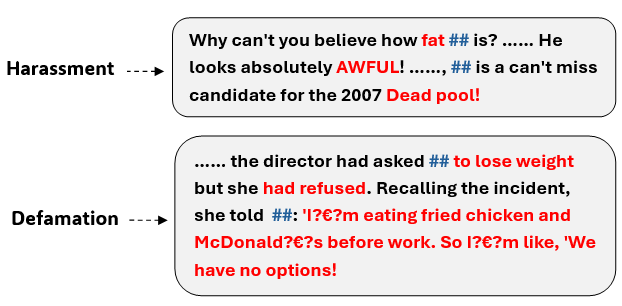}
  \caption{Examples of celebrity cyberbullying, with key terms highlighted and names anonymised as \#\#.}
  \label{fig:samples}
\end{figure}

Cyberbullying is a serious global issue, which is a specific form of bullying that happens within online environments \cite{Smith2008}.
Cyberbullying manifests itself in diverse forms, such as harassment (insults or threats), spreading rumours, impersonation, outing and trickery (sharing someone's confidential information) or exclusion (e.g. from activities) \cite{peled2019cyberbullying}. Despite the multifaceted nature of cyberbullying, existing research \cite{Agrawal2018,muneer2020comparative,kim2021human,yi2023session} has primarily focused on direct forms such as harassment and flaming, with limited exploration of indirect forms like denigration, which is an obstacle for addressing e.g. celebrity cyberbullying. Recent studies demonstrate that cyberbullying targeting influencers and celebrities has become commonplace \cite{takano2024online}. Furthermore, findings from \citet{ouvrein2018online} suggest that some adolescents do not perceive negative comments against celebrities as cyberbullying, but rather as legitimate, personal opinions. Nonetheless, celebrities often suffer serious consequences, including alcohol and drug addictions, self-blame, and depression. Figure \ref{fig:samples} illustrates two distinct forms of cyberbullying towards celebrities, which involve repeated aggressive (direct) and defamatory (indirect) behaviour.

In this work, we investigate multi-class celebrity cyberbullying detection, tackling three key bottlenecks in existing research: \textit{B1) Lack of suitable datasets}: previous attempts relying on offensive language as a proxy for cyberbullying data collection fail to capture indirect incidents of cyberbullying; \textit{B2) Heterogeneous classes:} existing research on multi-class classification in cyberbullying primarily focuses on discerning the severity levels of direct forms, overlooking the detection of distinct types such as harassment and defamation; \textit{B3) Dearth of research in low-resource settings:} Transformer-based models are widely used in advanced cyberbullying detection research, but there is a notable scarcity of research in the field targeting resource-poor environments, even though these environments are the most common context in cyberbullying studies.


\textbf{Proposed Approach:} 
To address these challenges, we initially 
 develop a new celebrity cyberbullying dataset called HDCyberbullying. This dataset contains two different forms of cyberbullying: harassment and defamation.
Then we propose an emotion-adaptive training framework (EAT) under low-resource settings to tackle B2$\&$B3. The approach is straightforward: given the limited data availability, our training emphasis shifts away from the cyberbullying detection dataset. Instead, we focus on acquiring knowledge in the emotion detection domain and subsequently transferring it to the cyberbullying detection task.

To evaluate the effectiveness and robustness of our method, we apply the method to nine transformer-based models: RoBERTa, Bert, DistilBert, Electra, XLnet, T5, Mpnet, Llama2 and Llama3. We conduct a series of experiments and quantitative analyses to understand why our approach can be successful across a broad range of models.

\textbf{Contributions:} To the best of our knowledge, our work is the first investigation into the detection of celebrity cyberbullying and multiple classifications of direct and indirect cyberbullying. Our contributions include:

\begin{compactitem}
 \item Our study underscores the transferability from the emotion domain to the domain of cyberbullying detection.
 \item To encourage diversity in cyberbullying research, we create the first dataset that reflects real-life cyberbullying scenarios involving celebrities.
 \item We propose the EAT framework to address heterogeneous class detection problems in cyberbullying within resource-poor environments.
 \item We quantitatively evaluate the efficacy of EAT with nine state-of-the-art models, along with qualitative evaluations to gain an in-depth understanding of why and how EAT achieves competitive and consistent results.
\end{compactitem}

\section{Related work}
Cyberbullying detection is generally defined as a binary classification task determining if an instance constitutes a case of cyberbullying or not. Cyberbullying detection research by \citet{Agrawal2018, Dadvar2018, Yuvaraj2021, Cheng2020} leveraged language models, including methods based on GloVe and Word2Vec in combination with deep learning methods such as BLSTM, CNN and attention mechanisms. Using Transformer-based models to identify cyberbullying through multiple user interactions demonstrates that transformer-based models can be strong, and competitive for cyberbullying detection \cite{gururangan2020don,yi2023session,yi2023learning}. Large language models(LLMs) have demonstrated robust comprehension capabilities \cite{liu2023pre} to human commands, particularly following advancements in reinforcement learning through Prompt Engineering 
\cite{ouyang2022training,alkhamissi2022review}. \citet{li2023hot} illustrate that ChatGPT exhibits high efficacy in detecting harmful textual content.

Existing research studies have primarily focused on direct forms of cyberbullying, such as harassment and flaming, with limited exploration of indirect cyberbullying, typified by denigration. Handling both overt and implicit forms of cyberbullying simultaneously remains a challenge. However, several works have shown awareness of this issue. For instance, \citet{saengprang2021cyberbullying,maftei2022using,scheithauer2021types} have all highlighted the prevalence of harassment and denigration in cyberbullying incidents involving public figures and adolescents. Some research efforts have specifically targeted denigration detection.  For example, \citet{sangwan2020denigration,sangwan2022d} employed various feature selection algorithms to enhance the relevance of features for classification of denigration. Furthermore, while emotions play a significant role in driving human behaviour, the exploration of their influence on improving cyberbullying detection is still in its early stages \cite{al2023cyberbullying}. In our study, we perform an in-depth empirical study on the strong connection between emotion and cyberbullying, demonstrating that models can effectively incorporate knowledge from emotion datasets to help identify indirect cyberbullying incidents.

Domain adaptation is the foundation of our approach. It is a type of transfer learning designed to adapt a model trained on a source domain so that it performs effectively on a target domain with a different data distribution. Domain adaptation in the context of cyberbullying detection is an emerging field initiated by \citet{agrawal2018deep}. They studied the performance of a zero-shot transfer learning approach on three different social platforms (Wikipedia, Twitter and Formspring), training and testing on different platforms. Their study underscored the complexity and challenges of the problem. To tackle this challenge, \citet{yi2022cyberbullying} proposed adversarial transformers by integrating unlabeled data from both source and target domains into a unified representation, thereby avoiding platform-specific training. Inspired by this research, we aim to achieve a common space of knowledge transfer between the emotion detection domain and the cyberbullying detection domain.

\section{HDCyberbullying}

In the past decade, NLP researchers made available several cyberbullying datasets for detection tasks \cite{yi2023session}, However, existing datasets 1) cover direct cyberbullying but lack coverage of indirect cyberbullying, and 2) do not simultaneously target celebrity harassment and defamation. To address these gaps, we compile HDCyberbullying. 

\subsection{The definition of celebrity cyberbullying}\label{defintion}

There is no established definition of celebrity cyberbullying. Based on existing studies \cite{takano2022online,karthika2022cyberbullying,EMILY2021}, we define celebrity cyberbullying as the incidents of repeated harassment and defamation of celebrities by name, where these two terms are defined as follows:

\begin{compactitem}
 \item {\textbf{Harassment:}} including \textbf{1) name-calling,} i.e. insulting someone by calling them rude names; \textbf{2) offensive abuse,} i.e. comments that cause mental pain through ``abusive appearance", ``abusive ability", ``abusive personality", etc.; \textbf{3) hateful comments,} i.e. offensive discourse towards targets based on race, religion, skin colour, sexual identity, gender identity, ethnicity, disability, or national identity.
 \item {\textbf{Defamation:}} speaking half-truths or lies.
\end{compactitem}

\subsection{Selection of comments}


According to the above definition, HDCyberbullying is exclusively designed for celebrity cyberbullying. By combining the harassment dataset \cite{su-cyberbullying} and the fake news dataset \cite{Perez-Rosas18Automatic}, both of which are relevant to celebrity disinformation. The harassment dataset \cite{su-cyberbullying} notes that many celebrities face backlash and encounter hateful and offensive comments. In our initial analysis, 70\% of the samples contained named comments, which is significantly different from the anonymous user attacks prevalent in other cyberbullying datasets. We used the Stanford NER Tagger \cite{StanfordNERTagger4.2.0} to identify harmful comments directed at celebrities by selecting those containing celebrities' and other people's names. We did not remove unfamiliar names, believing this would not affect the data quality.


The Defamation Dataset \cite{Perez-Rosas18Automatic} focuses on actors, singers, socialites, and politicians, and is collected from web sources to identify naturally occurring false content. The data collection was paired, with one article being legitimate and the other being false. To determine whether a celebrity news article is legitimate or not, claims made in the article were evaluated through gossip-checking websites and cross-referenced with information from other entertainment news sources. Using this method, 500 news articles with an even distribution of fake and legitimate news were collected.

\subsection{Annotation}

These original datasets already contain labels related to cyberbullying. For example, the Harassment dataset \cite{su-cyberbullying} includes six categories: `Malignant', `Highly malignant', `Rude', `Threat', `Abuse', and `Loathe', which are combined into a single `harassment' tag. Therefore, all samples in the HDCyberbullying dataset fall into one of three categories:\textbf{harassment, defamation, and non-cyberbullying}. 
We manually verify that each tag meets our definition above. Table \ref{tab:anotation} illustrates examples of annotation.

\begin{table*}[h]
\centering
\scalebox{0.94}{
\centering
  \label{tab:datasets}
  \begin{tabular}{c|c|c}
    \toprule
     Annotation& Posts         & Tag                 \\

        \midrule
    Offensive abuse &  ... my acctions on \textcolor{blue}{\#\#} abuse userpage ... i suspect he is a \textcolor{red}...{sockpuppet}  &1                       \\
   \midrule
   Hateful  comments& ...to call \textcolor{blue}{\#\#} a \textcolor{red}{moron} which is what he is...& 1              \\
  
     \midrule
     Name calling&...Now,\textcolor{blue}{\#\#}...the typical \textcolor{red}{right-wing}...&1\\
   
\midrule
    Defamation&...\textcolor{blue}{\#\#}: \textcolor{red}{Relapse Fears} \textcolor{red}{Over Showbiz Deaths}...&2\\
  
    \midrule
     No-Cyberbullying& ... getting more \textcolor{blue}{\#\#} pics if they're available...&0\\
    
    \bottomrule

  \end{tabular}
}
  \caption{Example annotation in HDCyberbullying, with key terms and names replaced by \#\#}.
  \label{tab:anotation}
\end{table*}






\subsection{Data statistics}
The dataset consists of 2,907 reviews, including 250 defamatory comments, 1,204 harassing comments, and 1,453 non-cyberbullying comments. There is a notable imbalance in both text length and category distribution, with defamatory data representing only 8\% of the total. In contrast, the average text length is 1985 words, and harassment data accounts for 41\%. The average length of harassing text is 455 words. Additionally, the diversity of language features and datasets poses challenges for multi-category tasks.


  

    

\section{Methodology}

This section defines the research problem and elucidates the theoretical underpinnings of our solution, thereby laying the foundation for understanding subsequent explanations of EAT.

\subsection{Problem definition}

\textbf{Definition 1: Cyberbullying detection.} A Multiclass classification task determines if each text in $T \in \{T_1, \ldots, T_n\}$ indicates a cyberbullying incident. i.e. $y \in \{0, 1,2\}$, where $y = 1$ means that a post refers to a case of harassment, $y = 2$ means that a post refers to a case of defamation, and $y = 0$ means that it is not a case of cyberbullying.

\textbf{Definition 2: Emotion detection.} A Multiclass classification task determines if each text in $E \in \{E_1, \ldots, E_n\}$ belongs to an emotion group related to cyberbullying incident. i.e. $y \in \{0, 1,2\}$, where $y = 1$ means that a post expresses one or more emotions 
indicative of harassment, $y = 2$ means that the emotions of a post most intensely towards defamation. When $y = 0$, it indicates that the detected emotions are unrelated to cyberbullying events.

\textbf{Definition 3: Zero-shot classification (ZSC).} In each experiment configuration, datasets belong to the training datasets $s$ or test datasets $t$, which leads to two different input spaces $X_s$ and $X_t$ where $X_s \neq X_t$, and they have different label space $Y_s \neq Y_t$. Moreover, the training dataset's data distribution is $P_s(x,y)$ and the test dataset distribution $P_t(x,y)$, which are different from each other, are both unknown and imbalanced. Zero-shot classification aims to learn a classifier for classifying testing instances belonging to the unseen label $Y_t$.

\textbf{Definition 4: Few-shot classification (FSC).} In each experiment configuration, $X_s \sim X_t$ and they have the same label space $Y_s  \equiv Y_t$. However, the number of training datasets is too small to help find a data distribution $P_s(x,y) \approx  P_t(x,y)$. Therefore, the few-shot classification algorithm is a parameterized optimal strategy for finding $P_t(x,y)$.

\textbf{Problem definition:} We assume that the emotion detection domain and the cyberbullying detection domain are related but from different distributions. However, a dataspace $D_e$ in emotion detection datasets can be found to transfer the knowledge from the emotion detection domain to the cyberbullying detection domain. Under this assumption, our task in the study is reducing the disparity across domains and training the emotion detection classifier $C_e$, which can be directly applied to instances from the cyberbullying detection domain (ZSC), on labelled emotion detection data. Or help improve performance of the cyberbullying classifier $C_t$ on low resource settings (FSC). 

\subsection{Theoretical Analysis}
\label{theory}

In this section, we will conduct a theoretical analysis of our method. Our analysis is based on \cite{Ben-David2010a} and \cite{Yosinski2014}'s foundational theoretical research on the adaptability of transfer learning. We explain how these theories support our approach to obtain substantially improved results over state-of-the-art models.

\textbf{Data space similarity:}
The disparity between the target and source domain can significantly affect the target task. We need high-quality source datasets to train a source classifier with a lower loss to maximise the effectiveness of the transmission. Hence, the initial step for EAT involves carefully selecting the source domain, and aiming to gauge the similarity of the cyberbullying detection dataspace to the emotion detection dataspace.

\textbf{Domain shift:} This part is to deal with the data shift ($P_s(x,y) \neq P_t(x,y)$) between the source domain and the target domain to make them similar ($P_s(x,y) \approx P_t(x,y)$). We can decompose the joint probability distribution as $p(x, y) = p(y)p(x|y)$ or $p(x, y) = p(x)p(y|x)$. The source and the target posteriors are arbitrary $P_s(x|y) \neq P_t(x|y)$. The problem becomes intractable when $P_s(y) \neq P_t(y)$. Thus we make the data distribution shift ($P_s(x) \approx P_t(x)$) first by selecting data with certain groups. Then concept shift makes $P_s(y|x) \approx P_t(y|x)$. This is achieved by grouping and mapping the emotion categories that are closely related to the cyberbullying category.

\textbf{A high-quality classifier:}
The extent to which knowledge from the source domain can be transferred to the target domain primarily depends on the hypothesis loss in the source domain. The performance of the model is contingent upon the training loss in both domains. If the combined loss is substantial, the model may struggle to perform effectively. Hence, we require a model structure that performs well in both domains. Transformer-based pre-trained models are trained on extensive and diverse corpora. Representations learned by such models demonstrate robust performance across numerous tasks, utilizing datasets of varying sizes and originating from diverse sources. Therefore, we extensively selected pre-trained models of different architectures as the knowledge containers for this study.

\subsection {Emotion adaptive training (EAT)}
\begin{figure}[tbh]
  \centering
  \includegraphics[width=\linewidth]
  {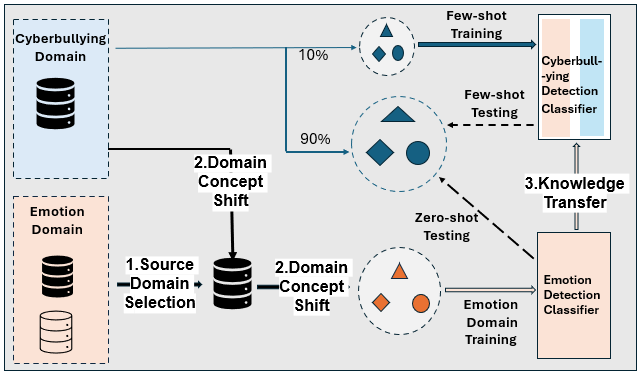}
  \caption{Architecture of EAT}
  \label{fig:structure}
\end{figure}

In this section, we detail our simple yet effective domain adaptation approach, which consists of three main phases: source domain selection, domain concept shift and knowledge transfer. The framework is depicted in Figure \ref{fig:structure}.

\textbf{Source domain selection:} This phase aims to perform the distribution shifts ($P_s(x) \approx P_t(x)$) by selecting a single emotion domain as the source domain from multiple potentially helpful emotion datasets. In the past decade, NLP researchers made several datasets available for language-based emotion classification for various domains and applications \cite{buechel2022emobank,scherer1994evidence,oberlander2020goodnewseveryone,liu2019dens,kajiwara2021wrime,demszky-etal-2020-goemotions}. To find a source domain relevant to the target domain and a high-quality dataset sufficient to train a high-quality emotion classifier, we need a large-scale, consistent labelled emotion dataset based on fine-grained taxonomies with proven high-quality annotations and from a single genre \cite{bostan-klinger-2018-analysis}. \cite{demszky-etal-2020-goemotions} is the largest manually annotated dataset of 58k English Reddit comments, labelled for 27 emotion categories and demonstrated the high quality of the annotations via
Principal Preserved Component Analysis.

\begin{figure}[tbh]
\centering
\includegraphics[scale=0.25]
  {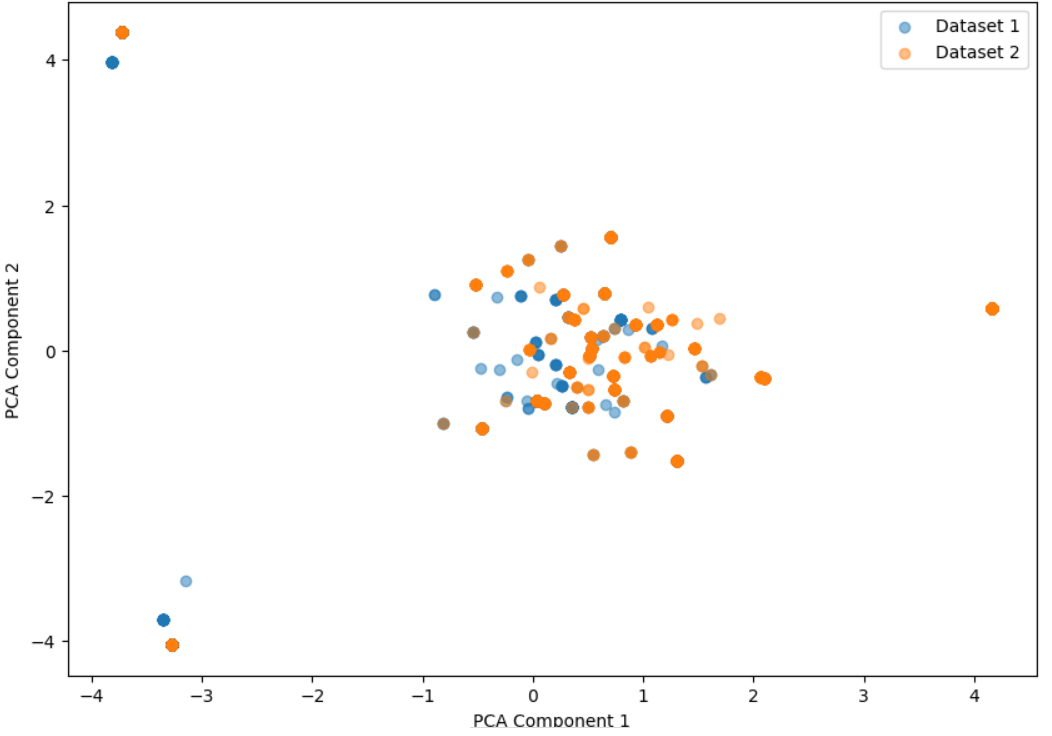}
  \centering
  \caption{Domain similarity. Orange: Emotion data; Blue: Cyberbullying data.}
  \label{fig:similarity}
\end{figure}

To intuitively assess the similarity between the emotion data domain and the cyberbullying data domain, we randomly selected 1000 samples from each domain. We then generated 768-dimensional RoBERTa embeddings and reduced them to 2 dimensions using PCA (Principal Component Analysis). PCA performs dimensionality reduction while preserving as much variance as possible. The resulting, average cosine similarity is 0.993. Figure \ref{fig:similarity} depicts the dataspace, where blue spots represent the cyberbullying dataset and orange spots represent the emotion dataset. The two domains overlap significantly. 

\begin{figure}[tbh]
  \centering
  \includegraphics[scale=0.63]
  {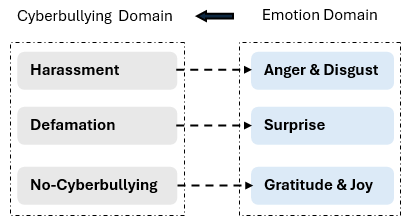}
  \caption{Mapping between Emotion and Cyberbullying domain.}
  \label{fig:map}
\end{figure}

\textbf{Domain concept shift:}
This phase ensures that $P_s(y|x) \approx P_t(y|x)$, which is achieved by grouping and mapping emotion categories that are closely related to the cyberbullying category. For instance, data instances labelled as `Anger' or `Disgust' in the emotion domain are grouped and assigned label 1, corresponding to the cyberbullying domain of `Harassment'.
Figure \ref{fig:map} illustrates this relationship, and is based on prior research identifying anger and disgust as powerful emotions shared by victims and bullies in both physical and electronic forms of bullying \cite{burgess2009cyberbullying,lonigro2015cyberbullying,wang2017trait,al2023cyberbullying}. Surprise is felt most intensely towards celebrity fake news \cite{tan2023application,liu2024emotion}. There are many choices for positive emotions in the emotion dataset. We chose `gratitude' and `joy' because the original authors achieved the best results for these two categories using the BERT-based emotion classification model \cite{demszky-etal-2020-goemotions}. This ensures the maximum potential for high-quality source classifiers from these two categories.

\textbf{Knowledge transfer:}
In this phase, we consider unsupervised domain adaptation for ZSC and semi-supervised domain adaptation for FSC. Unsupervised domain adaptation fine-tuning pre-trained models on labelled emotion datasets without requiring access to the cyberbullying dataset. Semi-supervised domain adaption based on the model trained by unsupervised domain adaptation do further training on a few cyberbullying data. How much knowledge can be transferred is decided by combining the loss of emotion classifier and the cyberbullying classifier.  

\section {Experiments}
This section conducts quantitative evaluation to assess the effectiveness of EAT, complemented by qualitative experiments aimed at providing a comprehensive understanding of the reasons behind EAT's consistent results.

\subsection {Transformer-based models}
 The different pre-training approaches are crucial for the success of downstream tasks. In the experiment, we tested nine transformer-based models with six types of pre-training approaches to examine the broad applicability of the EDA approach.

 \textbf{Masked:}
This approach is to mask some tokens in the input sequence, and the model learns to predict the masked tokens based on the surrounding context which has been shown to have significant advantages for text classification. We test three masked pre-trained models, namely BERT \cite{devlin2018bert}, Roberta \cite{liu2019roberta} and Distilbert \cite{sanh2019distilbert}.

\textbf{Autoregressive:}
XLnet \cite{yang2019xlnet} considering the dependence of masked tokens which missing in the masked approach, based on autoregressive language modelling.

\textbf{Masked+Permuted:}
Mpnet \cite{song2020mpnet} inherits the advantages of BERT and XLNet, and leverages the full position information to reduce the discrepancy between pre-training and fine-tuning by combining masked and permuted approaches in a view.

\textbf{Replacing:}
Electra \cite{clark2020electra} replacing some input tokens with sampled from a small generator network instead of masking the input.

\textbf{Text-text:} Text-to-text approach converts text processing problems into a universal text generation structure. T5 \cite{raffel2020exploring} will be examined in this experiment as a representative model of this approach.

\textbf{LLMs:} Large language models trained on a diverse and extensive dataset composed of publicly available text data. This data includes a wide range of domains, such as books, websites, and other forms of written content, to ensure the model captures a broad spectrum of language. Llama2 and Llama3, which utilize this technique along with extensive human preference data, demonstrate impressive performance on many NLP tasks \cite{touvron2023llama}.

\subsection{Settings}

All pre-trained models are based on Hugging Face's Enterprise Hub. To ensure a fair comparison, all models, except for the Llama2 and Llama3 models, use SimpleTransformer wrapper \cite{rajapakse2019simpletransformers}. The training hyperparameters recommended by \cite{Sun2019a} are as follows: batch size of 32, learning rate (Adam) of $4e^{-5}$, number of epochs set to 4, and a token length of 400, all trained on an L4 GPU.

\subsection{Evaluation}
We use three widely adopted evaluation metrics for each class, namely recall, precision and micro-F1. Additionally, we also report macro-averaged recall, precision, and F1 across all classes.

For the baseline, we fine-tune the pre-trained LLM on {$10\%$} of the data (291 samples) and then test it on the remaining {$90\%$} (2,524 samples). To evaluate the extent to which EAT can help pre-trained models improve generalization, we extracted 6 subsets of training data of different sizes: 72, 140, 210, 400, 700, 1300. Table \ref{tab:statistics} shows these training settings of each class and the number of testing data.

\begin{table}[h]
  \scalebox{0.65}{
  
  \begin{tabular}{c|c|c|c|c|c}
    \toprule
     & {\#}Harassment &{\#}Defamation&{\#}No-Cyberbullying&{\#}Emotion&{\#}Testing                \\
        \midrule
   Baseline  &  119 &20&152&0&2,615                      \\
       
    Zero-shot  & 0 & 0&0 &3700&2,615                \\  
   
    Few-shot& 119&20&152&3700&2,615    \\    
    
    72&29&5&38&3700&2,834\\
  
    140&57&10&73&3700&2,766\\
    
    210&86&14&110&3700&2,696\\
   
    400&164&27&209&3700&2,506\\
   
    700&286&48&366&3700&2,206\\
    
    1300&531&89&680&3700&1,606\\
    \bottomrule
  \end{tabular}
  }
  \caption{Data settings for evaluation.}
  \label{tab:statistics}
\end{table} In the zero-shot experiment, the training dataset is exclusively from the emotion detection domain, with no visibility of the cyberbullying detection dataset. In the few-shot experiment, we use the same fine-tuning and test data as the baseline. Each model is run 5 times to report average performance.

\subsection {Results}

\begin{table}[!htbp]
\scalebox{0.70}{   
  \begin{tabular}{c|c||c|c|c||c|c|c||c|c|c}
  \toprule
 \multicolumn{2}{c||}{}&\multicolumn{3}{c||}{Baseline}&\multicolumn{3}{c||}{Zero-shot (EAT)}&\multicolumn{3}{c}{Few-shot (EAT)}\\
     \midrule
Model&&P&R&F1&P&R&F1&P&R&F1\\
    \midrule

 RoBerta&H  &0.81& 0.91&0.85&0.75&0.94&0.83&
    0.84&0.95&\textbf{0.89}\\
  
    \_base&D&0&0&0&0.18&0.44&0.25&0.81&0.72 &\textbf{0.76}\\
    &A &0.58&0.55&0.56&0.60&0.61&0.57&0.85&0.83&\textbf{0.84}\\
     \midrule
Bert&H&0.85& 0.85&0.85&0.71&0.93&0.80&0.86&0.92&\textbf{0.89}\\
   
    \_base&D &0&0&0&0.17&0.10&0.12&0.73&0.66&\textbf{0.69} \\
    &A&0.57&0.52&0.54&0.56&0.56&0.55&0.82&0.81&\textbf{0.81}\\
   \midrule

Distilbert&H &0.87& 0.86&0.87&0.69&0.93&0.79&0.86&0.92&\textbf{0.89}\\
   
    \_base&D&0&0&0&0.15&0.19&0.17&0.80&0.43&\textbf{0.56} \\
    &A &0.54&0.59&0.56&0.56&0.56&0.55&0.84&0.74&\textbf{0.77}\\
   \midrule
 Mpnet&H&0.86& 0.91&0.88&0.82&0.92&0.87&0.87&0.94&\textbf{0.90}\\
   
    \_base&D &0&\t0&0&0.17&0.25&0.20&0.77&0.71&\textbf{0.74}\\
    &A &0.55&0.60&0.57&0.60&0.61&0.60&0.84&0.83&\textbf{0.84}\\
   \midrule

  Electra&H&0.90& 0.37&0.52&0.67&0.90&0.70&0.86&0.89&\textbf{0.87}\\
   
    \_small&D &0&\t0&0&0&0&0&0.80&0.37&\textbf{0.51}\\
    &A &0.49&0.44&0.42&0.45&0.48&0.44&0.82&0.71&\textbf{0.74}\\
   \midrule
XLnet&H   &0.87& 0.91&0.89&0.73&0.92&0.82&0.87&0.91&\textbf{0.89}\\
   
    \_base&D &0.79&0.31&0.44&0.09&0.11&0.10&0.71&0.72&\textbf{0.67}\\
    &A &0.82&0.69&0.72&0.53&0.53&0.52&0.82&0.80&\textbf{0.81}\\
   \midrule

T5&H &0.92& 0.54&0.68&0.65&0.90&0.75&0.83&0.86&\textbf{0.84}\\
   
    \_base&D &0.49&0.10&0.17&0.18&0.26&0.21&0.30&0.97&\textbf{0.45}\\
    &A &0.68&0.53&0.54&0.54&0.54&0.50&0.67&0.78&\textbf{0.65}\\
 \midrule
   
Llama3&H &0.71& 0.70&0.72&0.91&0.85&0.87&0.90&0.91&\textbf{0.91}\\
   
    \_8B&D &0.66&0.58&0.52&0.75&0.84&0.56&0.72&0.67&\textbf{0.69} \\
    &A &0.69&0.67&0.65&0.79&0.73&0.74&0.83&0.81&\textbf{0.82}\\
    \midrule
Llama2&H &0.63& 0.62&0.63&0.75&0.89&0.82&0.86&0.89&\textbf{0.87}\\
   
    \_7B&D &0.39&0.65&0.49&0.14&0.32&0.19&0.60&0.67&\textbf{0.63} \\
    &A &0.56&0.61&0.57&0.57&0.57&0.55&0.78&0.80&\textbf{0.78}\\
    \bottomrule
\multicolumn{2}{c||}{Overall}
&0.61&0.58&0.57&0.58&0.58&0.56&\textbf{0.80}&\textbf{0.79}&\textbf{0.78}\\ 

    \bottomrule
  \end{tabular}}
  \caption{ A comparison of models' performance. The best results and models are highlighted in bold. P: Precision; R: Recall; F1: Micro F1;H: harassment; D: defamation; A: Macro-averaging; Overall: Average of A.}
   \label{tab:results}
\end{table}

Table \ref{tab:results} show that our proposed method EAT can effectively improve the average macro F1, precision and recall by 20\%. We report results for both classes: harassment(H) and defamation(D) with Precision, Recall and F1.  The following are observed: 1) In case of harsh data imbalance (20 defamation samples for training), our EAT approach wasn't biased to the majority class, and learning meaningful patterns and relationships made the model generalize well when encountering unseen instances from these classes. 2) Mask-based models in small data settings (291 training samples) without EAT fine-tuning are ineffective at identifying defamation. These models use bidirectional context, which may not always capture dependencies, especially in complex language constructs. Moreover, due to the inherent constraints of mask-based models, they perform well with shorter to moderate-length sequences, given their training objective. The average text length of a defamation message is 1985, which poses a challenge. In contrast, XLnet considers the dependence on masked tokens which missing in the masked approach show potential generalizability. 3) T5 and LLaMA models demonstrate excellent scalability, particularly for long text tasks, despite differences in their architectures and training objectives. T5 employs an encoder-decoder architecture and is pre-trained with a span-corruption objective, where random spans of text are masked, and the model learns to reconstruct them. This approach allows T5 to effectively handle context over longer spans of text. In contrast, LLama models use a decoder-only architecture designed for scalability. With large model sizes that capture extensive context, LLama can be fine-tuned on specific tasks or datasets that involve long text. 4) Mpnet is one of the best-performing models across zero-shot and few-shot tasks. Mpnet leverages the dependency among predicted tokens and auxiliary position information to reduce the disparity between the training domain and fine-tuning domain can be demonstrated effectively in the study.

\begin{figure}[tbh]
  \centering
  \includegraphics[width=\linewidth]
  {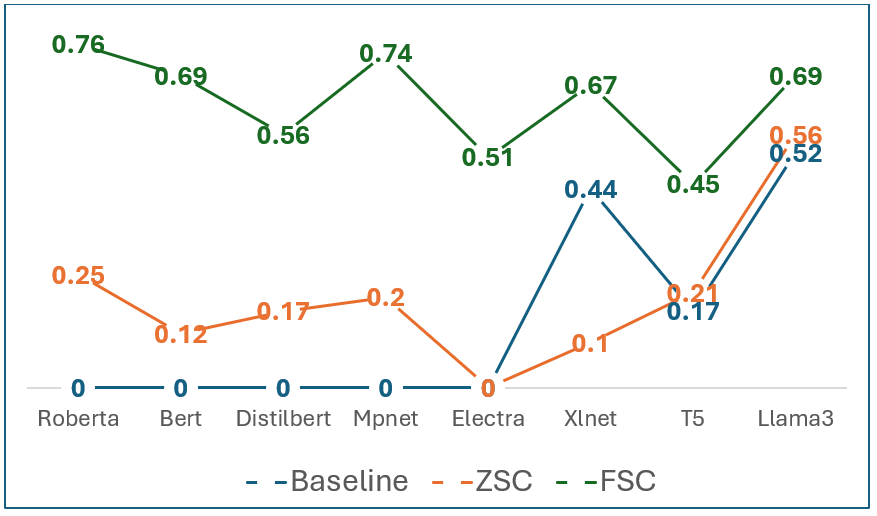}
  \caption{Performance of EAT in the defamation detection task.}
  \label{fig:defamation}
\end{figure}

In Figure \ref{fig:defamation}, we can visually compare the performance of EAT on the defamation detection task. RoBERTa, Bert, DistilBert, Mpnet and Electra overlook the minority class with zero on the F1 score. while EAT helps find robust presentations and data patterns leading to optimal performance.

\subsection {Error Analysis}
\begin{figure}[tbh]
  \centering
  \includegraphics[width=\linewidth] {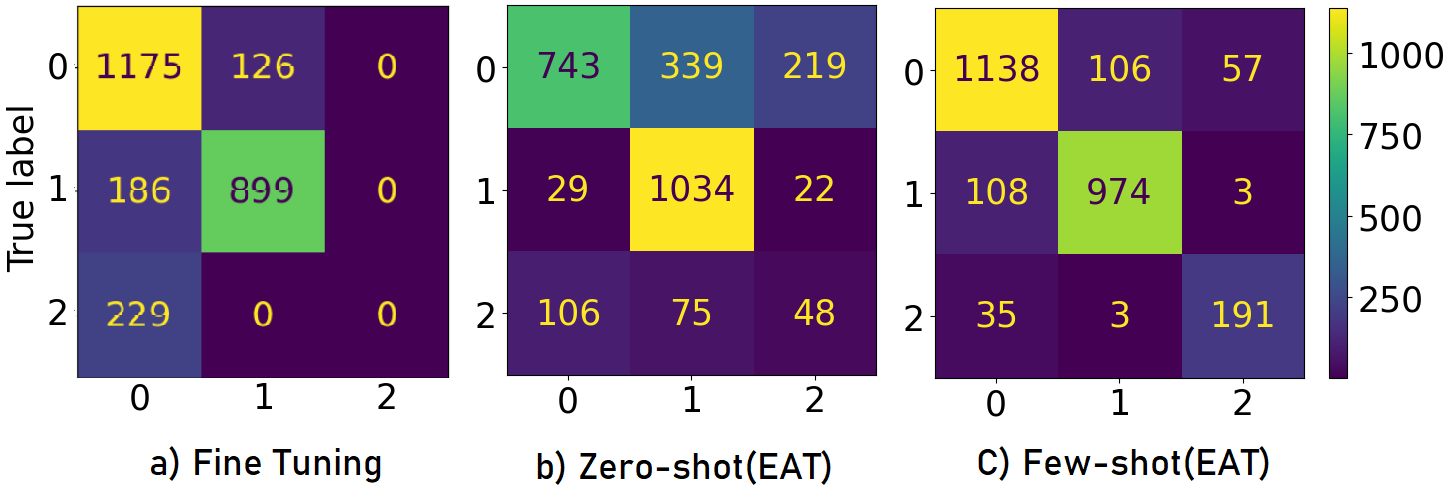}

  \caption{Confusion matrix of EAT. 0: No-cyberbullying; 1: Harassment; 2: Defamation.}
  \label{fig:confusion}
\end{figure}

To gain insights into the models’ behaviours, we look at the confusion matrix and manually analyse the classification error samples of the best-performing approaches, i.e. RoBerta\_base+EAT. Figure \ref{fig:confusion}a) illustrates that the RoBERTa\_base model is capable of detecting direct cyberbullying but struggles to differentiate indirect cyberbullying messages from non-bullying ones in low-resource fine-tuning. Defamation, like other forms of indirect cyberbullying, is ambiguous in its potential to be bullying. 

Figure \ref{fig:confusion}b) shows the results when the model is trained solely on emotion datasets. The model exhibits high accuracy in classifying harassment, indicating that anger and disgust are strongly associated with harassment events. It is also observed that around 60\% of non-cyberbullying events express emotions such as joy and gratitude. Interestingly, anger or disgust are also related to non-cyberbullying events. Past studies have found that surprise is most intensely associated with celebrity fake news \cite{tan2023application,liu2024emotion}. Our experimental results suggest that the relationships between emotions and celebrity defamation incidents are more complex than previously understood. Figure \ref{fig:confusion}c) illustrates that, except for a slight drop in the accuracy of detecting harassment from 95\% to 89\%, the EAT model's performance improves significantly for the other two classes.
 
Among the misclassified harassment examples, we find that some messages have an overall tone that is confrontational and assertive, with multiple elements suggesting the speaker is angry and with aggressive attitude. However, these messages are not instances of harassment. From the context, they appear more like defensive statements, as illustrated by sample 1 in Table \ref{tab:misclass}. This indicates that contextual knowledge is essential for the model to accurately distinguish such cases. The observation of inconsistency in sample 2 in Table \ref{tab:misclass} and the resulting question are typical reactions to something surprising. However, the statement appears to be a factual observation about the difference between the trailer and the final film and does not contain any false information. Factual news can evoke surprise due to the unexpected, unusual, or unprecedented nature of the events or information being reported \cite{qiu2024curiosity}.

Interestingly, our emotion classification model, EAT, classifies 70\% of defamation as joy or gratitude rather than the surprise typically identified in past studies. For example, sample 3 in Table \ref{tab:misclass} contrasts with traditional cyberbullying research, which often associates such behaviour with a range of negative emotions from both the perpetrator and the victim. In defamation, positive context and emotions can still lead to negative consequences. Therefore, more nuanced emotional indicators need to be explored in future research.

Although anger or disgust are the primary emotions associated with harassment events, some misclassified examples include messages where the perpetrator or victim is not explicitly named. For instance, sample 4 in Table \ref{tab:misclass} is not predicted as harassment. The meaning is obscured by the use of ambiguous expressions, such as the question ``Are you Jewish?" which reflects curiosity about the other person's background. The follow-up question, ``How do you see any resemblance between the Holocaust and Armenian deportations?" suggests an attempt to understand the other person's perspective or reasoning. Additionally, the phrase ``This may sound like a stupid question, but don't take it the wrong way" indicates an awareness that the question might be perceived as offensive or inappropriate. While these elements show a level of sensitivity and caution, the underlying racist implication remains.

Although the above analysis does not encompass all instances of model misclassification, it elucidates several underlying factors contributing to this phenomenon. Notably, the current intersection of research between the fields of cyberbullying detection and emotion analysis remains incomplete, particularly concerning indirect cyberbullying events. This gap underscores the need for further exploration and research, which may yield more effective detection models. We propose to address these limitations in future studies.

\begin{table*}[h]
\scalebox{0.85}{ 
\begin{tabular}{l|p{10cm}|l|l|p{2cm}}
    \toprule
    Index & Posts & Emotion & Cyberbullying & Prediction \\
    \midrule
    1 & Do not harass other editors with no basis. 
        You have posted a baseless warning/threat on my user page. 
        Do not post any more things on my page. 
        Otherwise, I will complain to the administrators, 
        or even \textcolor{blue}{\#\#} if I have to? 
      & Anger & No & Harassment \\
    \midrule
    2 & Trailer vs. Final Film: I noticed that in the trailer 
        for the film, \textcolor{blue}{\#\#} had two hamster butlers, 
        while in the actual film, he's the only pet living in the house. 
        What's up with that? 
      & Surprise & No & Defamation \\
    \midrule
    3 & \textcolor{blue}{\#\#}’s New Puppy Helping Her Heal With 
        ‘Love \& Support’ After Kidney Transplant. 
        \textcolor{blue}{\#\#}’s pooch is making her happier than ever 
        and he’s helping her heal with ‘love and support’ after her 
        kidney transplant, an expert revealed to HL EXCLUSIVELY. 
        Find out why the dog’s affection is so magical! 
      & Joy & Defamation & No \\
    \midrule
    4 & \textcolor{blue}{\#\#}, this may sound like a stupid question, 
        but don’t get it wrong. Are you Jewish? If so, how are you really 
        thinking there is any resemblance between the Holocaust and 
        Armenian deportations? 
      & Curiosity & Harassment & No \\
    \bottomrule
\end{tabular}
}
\caption{False Positive and False Negative examples of EAT, with names replaced by \#\#.}
\label{tab:misclass}
\end{table*}

\subsection{Insights from EAT}

\subsubsection {Training loss}
\begin{figure}[tbh]
  \centering
  \includegraphics[width=\linewidth]
  {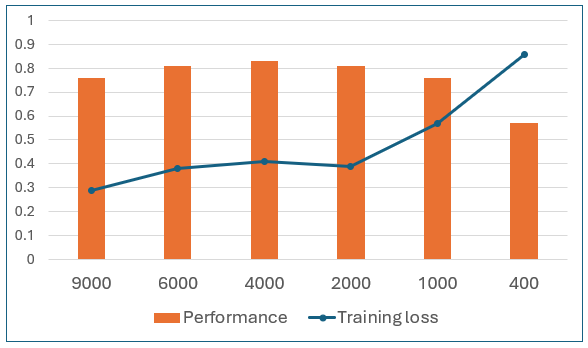}
  \caption{Training loss. The x-axis represents the size of the emotion training data and the y-axis represents the loss of training. }
  \label{fig:classifier_loss}
\end{figure}

Based on the theory discussion in Section \ref{theory}, we need to train a source classifier with a lower loss to maximise the effectiveness of transmission. we will specify the discussion to ``how many datasets do we need to train a good source classifier to help the target classification task". Figure \ref{fig:classifier_loss} illustrates the relation between training loss and model performance by using Mpnet+EAT. In general, there is an inverse relationship between emotion classifier training loss and model performance. However, when we continue to feed data, the training loss still decreases and the performance drops as well. Moreover, we observe that, when increasing data input from 2000 to 4000, the training loss is slightly increased. Our conjecture is that, when more data is input, the data diversity increases but the performance is the highest.  In conclusion, reducing source classifier training loss is general, but need to consider more factors when a lack of target samples is present. For example, the boundary of source domain data, the quality of data, model structure, etc.

\subsubsection{Knowledge transfer}
\begin{figure}[tbh]
  \centering
  \includegraphics[width=\linewidth]
  {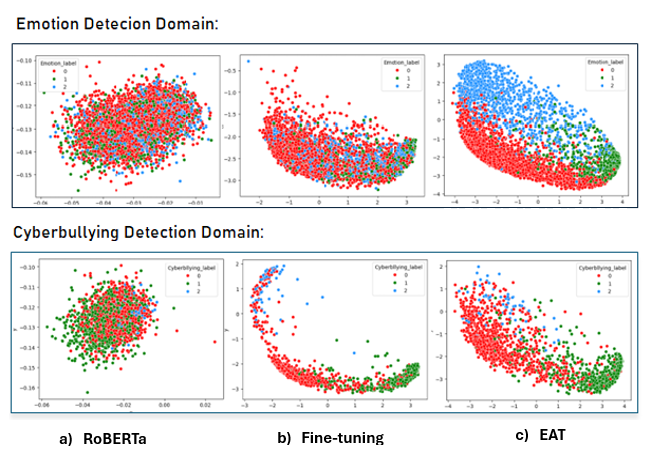}
  \caption{EAT t-SNE. Red: No-Cyberbullying samples, emotion (Gratitude and Joy) samples; Green: Harassment samples, emotion samples (Anger and disgusting); Blue: Defamation samples, emotion samples (Surprise).}
  \label{fig:emotion_tsne}
\end{figure}

To visualize the transfer capabilities of EAT, Figure \ref{fig:emotion_tsne} displays the distribution of embeddings generated by three RoBERTa-based models on two domains. We can observe a robust correlation between emotions and cyberbullying and how our emotion-adaptive training framework effectively helps data clustering in both domains. Figure \ref{fig:emotion_tsne}b) indicates that despite RoBERTa's pre-training corpus being sourced from diverse origins, it still encounters challenges when fine-tuning for cyberbullying detection tasks in suboptimal settings. However, Figure \ref{fig:emotion_tsne}c) illustrates that our method EAT efficiently transfers class knowledge from emotion detection to aid in the cyberbullying detection task. A reason is our selection of an emotion detection domain similar to the cyberbullying detection domain, as depicted in Figure \ref{fig:emotion_tsne}a)

\begin{figure}[tbh]
  \centering
  \includegraphics[width=\linewidth]
  {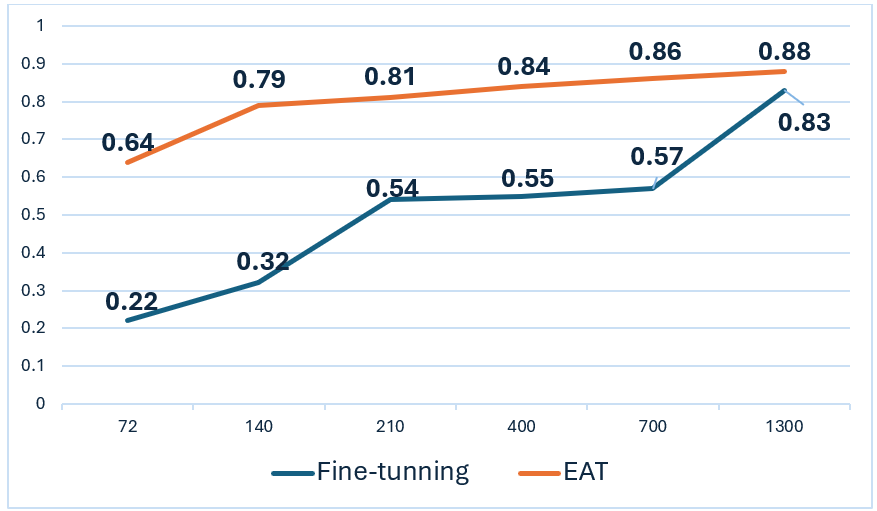}
  \caption{Capacity of transfer. The x-axis represents the size of the training data and the y-axis represents F1.}
  \label{fig:capacity}
\end{figure}

To evaluate the extent to which EAT can help models improve generalization, we extract 6 subsets of training data of different sizes, respectively 72, 140, 210, 400, 700, 1300, which are shown in Table \ref{tab:statistics} about these training settings of each class and the number of testing data. From Figure \ref{fig:capacity}, we observe that the less data available, the more beneficial EAT becomes. When the training data is approximately 50\% of the total data (1300 samples), the performance of the model without EAT fine-tuning is comparable to that of the EAT model, which requires only 10\% (210 samples) of the data. EAT enables models to extract more transferable features or representations, enhancing generalization to unseen data.

\subsubsection{Emotion reference}

\begin{figure}[tbh]
  \centering
  \includegraphics[width=\linewidth]{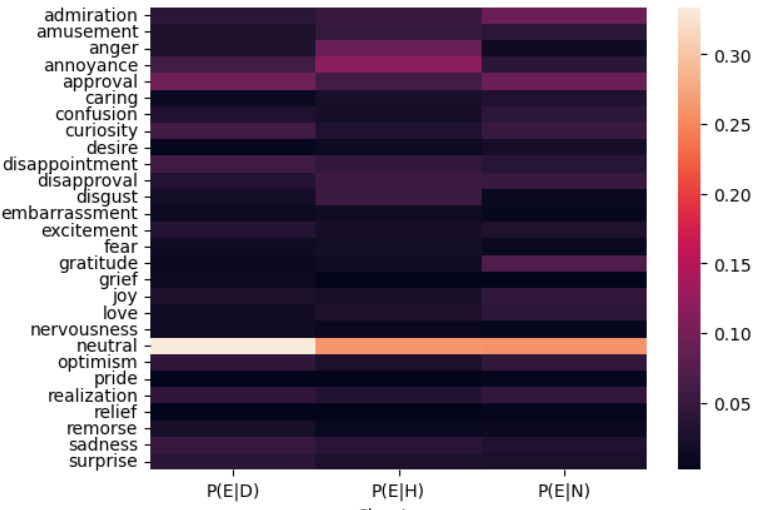}
  \caption{The likelihood of a certain emotion identifying given a class of data in the cyberbullying domain; E: emotion; H: Harassment; D: Defamation; N: No-cyberbullying. }
  \label{fig:heatmap}
\end{figure}

One of the reasons why EAT can be successful is that we have applied concept shift  $P_s(y|x) \approx P_t(y|x)$ to make $P_s(x,y) \approx P_t(x,y)$. We choose the emotion label according to past research, intuition and preliminary experiments. To gain insight into the relationship between cyberbullying and emotion, we calculate the likelihood of 28 emotions.  From Figure \ref{fig:heatmap}, we can observe that the emotions we choose have a high likelihood but not the maximum in each group. Neutral is the most effective for all three types of cyberbullying
affecting emotion labels. This shows that although the two data are similar, there is no one-to-one correspondence. In addition, for $P(E|H)$,  annoyance has a high likelihood value like anger. For $P(E|D)$ we choose ``surprise" according to the past research, but ``approval" is more like the emotion the rumour wants to generate, and ``admiration" looks more positive attitude related to the no-cyberbullying event.  At the same time, we notice all these emotions with high likelihood values didn't belong to \cite{ekman1992there}  or \cite{plutchik1980general}, usually used for emotion detection. So these findings call for further research into the relationship between cyberbullying and emotion.

\section {Conclusion}

We investigate a wide range of transformer-based pre-trained models and find that they struggle in low-resource settings for multi-cyberbullying detection tasks, especially with indirect cyberbullying like defamation, as summarized in Table \ref{tab:results}. We propose a novel, straightforward yet powerful domain adaptation framework, EAT, to tackle the above issue. Our experiments with EAT reveal that it may be valuable to complement work on ever-larger language models with parallel efforts to identify and use domain- and task-relevant emotion indicators to generalize models. The methods we studied are general enough to be applied to different structure transformer-based models. Our work points to numerous future directions, such as more comprehensive cyberbullying forms detection, better emotion indicator studies, efficient adaptation of large pre-trained language models to distant domains, and how to fine-tune language models after adaptation.

\section{Limitations}

Our proposed EAT model demonstrates state-of-the-art performance in this study. However, our work is not without limitations. Most importantly, the lack of more and broader datasets limits the detection of various forms of cyberbullying beyond harassment and defamation, such as outings and frapping, which are more complicated forms. Furthermore, although our study demonstrates the transferability of emotion domain data to cyberbullying data, the selection of more effective source domain data and emotion indicators requires further research.

\section{Ethics Statement}
Our research aims to support all individuals equally by curbing incidents of celebrity cyberbullying, particularly on social media, without intentionally discriminating against or causing harm to any vulnerable group. Our data comes from publicly available datasets and contains no personal information about each user. However, our research is not without risks, as adversaries involved in cyberbullying incidents could potentially use our findings for nefarious purposes, such as learning how to evade detection. This is not the intended use of our study.

\bibliography{Emotion_Cyberbullying}

\appendix



\end{document}